\documentclass{article}
\usepackage{ijcai19}
\usepackage[english]{babel}

\usepackage{hyperref}
\usepackage{times}
\usepackage{soul}
\usepackage{url}
\usepackage[utf8]{inputenc}
\usepackage[small]{caption}
\usepackage{graphicx}
\usepackage{amsmath}
\usepackage{booktabs}
\usepackage{graphicx}
\usepackage{authblk}
\usepackage{algorithm}
\usepackage[noend]{algpseudocode}
\title{Comparison of Privacy-Preserving Distributed Deep Learning Methods in Healthcare}

\author[1]{\textbf{Manish Gawali}}
\author[2]{\textbf{Arvind C S}}
\author[1]{\textbf{Shriya Suryavanshi}}
\author[1]{\textbf{Harshit Madaan}}
\author[1]{\textbf{Ashrika Gaikwad}}
\author[2]{\textbf{Bhanu Prakash KN}}
\author[1]{\textbf{Viraj Kulkarni}}
\author[1]{\textbf{Aniruddha Pant}}
\affil[1]{DeepTek Inc}
\affil[2]{Singapore Bioimaging Consortium - A*Star}

\begin{document}
\maketitle

\begin{abstract}
In this paper, we compare three privacy-preserving distributed
learning techniques: federated learning, split learning, and SplitFed.  We use these techniques to develop binary classification models for detecting tuberculosis from chest X-rays and compare them in terms of classification performance, communication and computational costs, and training time. We propose a novel distributed learning architecture called SplitFedv3, which performs better than split learning and SplitFedv2 in our experiments. We also propose alternate mini-batch training, a new training technique for split learning, that performs better than alternate client training, where clients take turns to train a model.

\end{abstract}

\section{Introduction}

There is a shortage of labeled data available in the healthcare domain, and even if it is available, healthcare data is commonly distributed and needs to be aggregated at a centralized storage site so that deep learning models can be trained. However, most of the healthcare centers and laws at the country level such as the General Data Protection Regulation (GDPR) and the Health Insurance Portability and Accountability Act (HIPAA) are rightfully protective of the data and do not allow free sharing of data across computer networks and national boundaries. Distributed learning methodologies solve this problem by enabling models to train using data from various healthcare centers without compromising the privacy of the data at these centers.

\subsection{Federated Learning}
Federated learning (FL) \cite{konevcny2016federated} \cite{FederatedLearningGoogleAI} \cite{mcmahan2017communication} is a distributed learning method that enables training of neural network models across multiple devices or servers without the need for movement of data. This is in contrast to centralized training where all the data samples from various data sources have to be collected at a centralized processing site. In FL, multiple federated rounds are performed to obtain a robust model. The workflow for one federated round (Figure 1) consists of the following steps: (i) pushing the global model from the main server to the clients (healthcare centers), (ii) training models on all healthcare center servers, and sending the local updates to the main server. (iii) The main server aggregates the updates received from the centers and upgrades the global model using federated averaging algorithm. This new global model is robust as it has learned from a large and diverse set of data\cite{roth2020federated}. Also, each healthcare center benefits as it can use a model which has also learned from some other healthcare center's data.

\begin{figure}
\includegraphics[width=\linewidth]{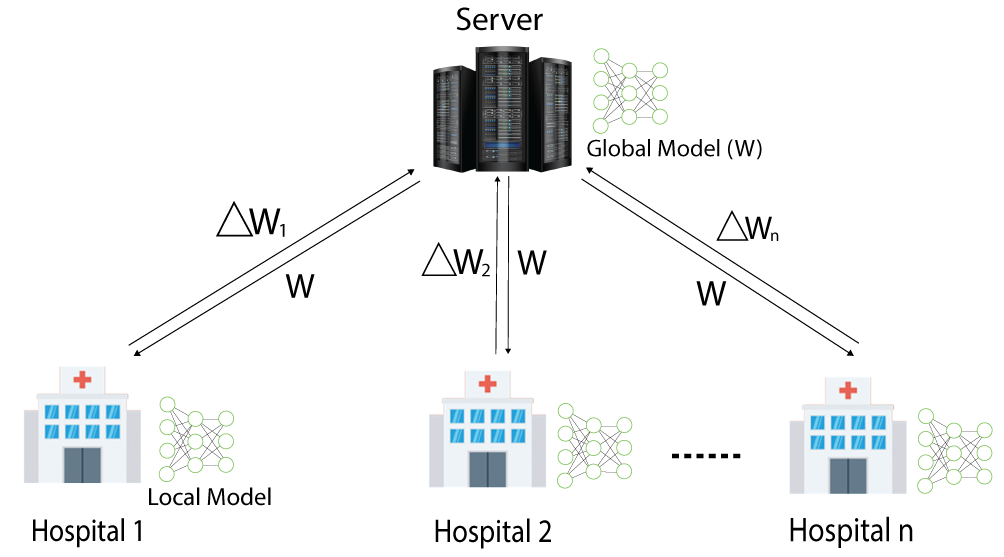}
\caption{Federated Learning - n hospitals collaborate to train a global model. The global model W, which is present at the server is transmitted to all hospitals, where each hospital trains a model with their own data. The server collects the updates $\Delta$W\textsubscript{i}'s from all hospitals and aggregates them to produce a single global update. This global update is used to tweak the global model W.}
\end{figure}

\subsection{Split Learning}
Split learning (SL) \cite{gupta2018distributed} consists of training a machine learning model across multiple hosts by splitting the model into multiple segments. In the simplest split learning configuration called label-sharing configuration in which labels for data are present on the server, each client (healthcare center) performs one step of forward propagation step till a particular layer called the cut layer\cite{gupta2018distributed} as shown in figure 2. The outputs at the cut layer are sent to the server, where the forward propagation is carried out on the rest of the network to generate predictions. The training loss is calculated at the server, using labels and predictions. A backpropagation step is performed on the network that is present at the server, i.e up to the cut layer. The gradients are sent back to the client so that a backpropagation step can be carried out on the first segment of the model. This process is repeated multiple times to obtain a final model. The server cannot access the raw local client data in the training process, thus preserving the privacy of the client.

\begin{figure}
\includegraphics[width=\linewidth]{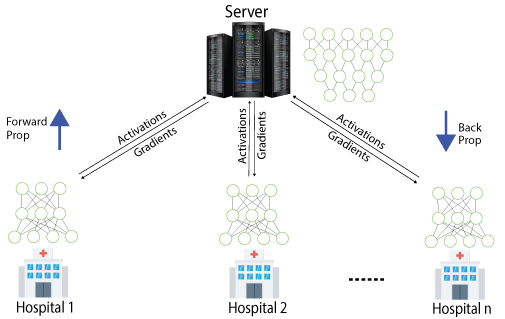}
\caption{Split Learning - The neural network architecture is segmented into two parts. The first segment resides at the hospital (client) and the second segment (common across all hospitals) resides on the server. n hospitals collaborate to train a global server-side model and n client-side models.}
\end{figure}

\subsection{SplitFed}
SplitFed learning (SFL) is a new decentralized machine learning methodology proposed by Thapa et al.\cite{thapa2020splitfed}, which combines the strengths of FL and SL. In the simplest configuration called the label sharing configuration, the entire neural network architecture is 'split' into two parts. Instead of training the client networks sequentially, Thapa et al. proposed training the client networks parallelly, which is a property drawn from FL. There are two variants of splitfed: SplitFedv1 (SFLv1) and SplitFedv2 (SFLv2). In SFLv1, clients perform a forward propagation step in parallel on their respective data and send the activations obtained at the cut layer to the main server. The main server performs forward propagation on the server-side network for all client activations in parallel. Subsequently, the server performs a backpropagation step and sends back the gradients to respective clients. At this time, the main server updates the server-side network using a weighted average of gradients obtained from backpropagation step. The clients perform a backpropagation step using the gradients obtained from the server and send the updates to fed server as shown in figure 3. Fed server averages the updates received from all clients and sends out a single update to all clients. The clients use this aggregate update to tweak their models. Therefore, the client and server-side networks are synchronized. In SFLv2, the training of the server-side network is sequential; i.e, clients perform forward propagation and backpropagation one by one sequentially. The client networks are synchronized at the end of each epoch by averaging all client updates at the fed server. 

\begin{figure}
\includegraphics[width=\linewidth]{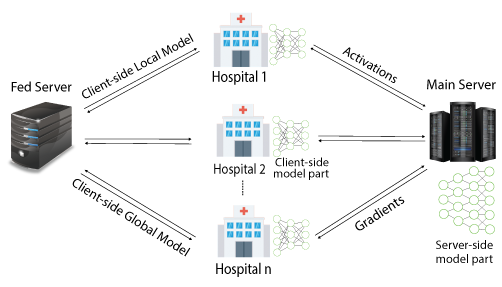}
\caption{SplitFed - The neural network architecture is segmented into two parts: client-side model and server-side model. n hospitals collaborate to train a global server-side  or n server-side models and global client-side or n client-side models depending upon the variant. Fed server averages updates from client-side models and main server averages updates for server-side models.}
\end{figure}

\section{Related Work}
Sheller et al.\cite{sheller2018multi} implemented federated learning in the medical domain for the first time. They demonstrated that U-Net models trained on the BraTS dataset using federated learning and models trained by traditional centralized method had similar dice scores. The concept of differential privacy was applied by Li et al.\cite{li2019privacy} for federated learning. Li et al. used a segmentation model for the BraTS dataset to show that incorporation of differential privacy slows down the convergence of the FL model.

Gupta et al.\cite{poirot2019split} introduced split learning and applied the U-shaped split configuration in the medical domain. They compared SL with two techniques,  centrally hosted and non-collaborative configuration, for two sets of problems: binary classification (fundus images) and multi-class classification (chest X-rays). With an increase in the number of clients, the performance of split learning remained stable, whereas the performance for the non-collaborative technique declined continuously.

Liu et al.\cite{liu2020experiments} used the federated learning framework for different deep learning architectures to detect COVID-19 using chest X-rays. Roth et al.\cite{roth2020federated} demonstrated that models trained on mammography data from multiple data sources using federated learning perform better than standalone models trained on data from a particular data source.

Prior works have compared distributed learning methods with centralized training but not with other distributed learning methods for application in the medical domain. In this comparative study, we evaluate the cost (in terms of classification performance, training time, communication, and computational costs) of using distributed learning in practice. Further, we contribute to this field by introducing a novel distributed learning architecture called SplitFedv3 (SFLv3) and a new training method called alternate mini-batch training. We implement these innovations and compare them with existing distributed learning techniques and training methods.

\section{Data and Methods}

This section describes the datasets and experimental setup for distributed learning methods.

\subsection{Data}

We obtained chest X-ray scans from five different sources. Three of these were private datasets, which we refer to as DT\textsubscript{1}, DT\textsubscript{2}, and DT\textsubscript{3}. The remaining two were publicly available research sets MIMIC\cite{johnson2019mimic} referred to as DT\textsubscript{4} and Padchest\cite{bustos2020padchest} referred to as DT\textsubscript{5}. A team of board-certified radiologists manually annotated these X-ray images using a custom built annotation tool. X-rays which showed indications of infiltrates, nodular shadows, cavitation, breakdown, lymph nodes, pleural effusion, bronchiectasis, fibrosis, scar, granuloma, nodule, pleural thickening, calcification, calcified lymph nodes, calcified pleural plaques were labelled as TB-suspect. Images which did not show these indications were labelled as TB-negative. In addition to these labels, the radiologist also drew polygon masks around the region of interest in which these manifestations were observed. Table 1 describes the dataset distribution and number of training, validation and test data taken from various sources. For each data source, the percentage of images belonging to the class TB-suspect (prevalence) in the training set is 50\%. The prevalence in the validation and test sets is 10\%. For experimentation, two different image resolution data was considered (i) 224x224 for densenet architecture (ii) 768x768 for U-Net architecture. 

\begin{table}[hbt!]
\centering
\resizebox{.48\textwidth}{!}{%
\begin{tabular}{ | c | c c c c c c | }
\hline
Data & DT\textsubscript{1} & DT\textsubscript{2} & DT\textsubscript{3} & DT\textsubscript{4} & DT\textsubscript{5} & Total\\ 
\hline
Train & 3772 & 1150 & 1816 & 880 & 1090 & 8708\\ 
Validation & 500 & 500 & 500 & 500 & 500 & 2500\\ 
Test & 500 & 500 & 500 & 500 & 500 &  2500\\ 
\hline
\end{tabular}
}
\caption{Distribution of TB CXR images}
\end{table}

\subsection{Topology and Neural Network Architectures}
The experimental network topology consists of one server and five clients, where each client has data from a single data source. The clients are virtual workers i.e they reside on the same machine as the server. We chose this topology as it is close to the practical setting where hospitals (clients) are likely to have non-I.I.D data.

All of our experiments were done using PySyft\cite{ryffel2018generic}. We performed two sets of experiments for classification by varying the model architecture. For the first set, we used DenseNet-121 architecture\cite{huang2017densely}. For the second set, we used the U-Net architecture\cite{ronneberger2015u} with Xception as the backbone. The U-Net architecture is traditionally used for segmentation problems, but we used it for a classification task by deriving probabilistic output from segmentation output. For both sets of experiments, we used binary cross-entropy as the loss function and the Adam optimizer\cite{kingma2014adam} with standard parameters ($\beta$\textsubscript{1} = 0.9 and $\beta$\textsubscript{2} = 0.999) and learning rate of  10\textsuperscript{-4}. The batch size was 64 in DenseNet experiments and 4 for U-Net experiments. DenseNet models were trained for 10 epochs, whereas U-Net models were trained for 5 epochs. These models were trained for the stated number of epochs as they converge within those number of epochs. We saved the model with the least validation loss on the validation set and evaluated it on the test set.
The specifications for the machine used for the experiments were 8 GB RAM, Ubuntu 18.04 OS, Tesla T4 16 GB GPU.

\subsection{Federated Learning Settings}
For federated learning models, we used federated averaging algorithm\cite{mcmahan2017communication} to update the global neural network model at the end of each federated round (epoch). We do not address the concept of differential privacy for the experiments.

\subsection{Split Learning Settings}
We experimented with two split learning configurations: the vanilla split learning/label sharing (LS) configuration and the U-shaped split-learning/ non-label sharing (NLS) configuration as shown in the figure 4. In the LS configuration, the input images remain with the clients and the labels go to the server, whereas in the NLS configuration, the input images and the labels are both present with the clients.

We trained the split learning model using the alternate client (AC) training and the alternate mini-batch (AM) training techniques. In alternate client training, the clients train their networks on their entire data sequentially, and the server network, which is common for all clients, updates sequentially as well. In alternate mini-batch training, a client updates its network on one mini-batch, after which the client next in order takes over. As the number of data samples can vary for each client, if some client finishes up with its mini-batches, then it has to wait until the next epoch starts, during which other clients can continue training on mini-batches sequentially. So, sequential updates on mini-batches distinguish the server-side training in alternate mini-batch training from the server-side training in alternate client training.

In the DenseNet experiments, the network was split such that first 4 layers are at the client end and the rest of the network is at the server for the label-sharing configuration. For the non-label sharing configuration, the last fully connected layer is present at the client-side in addition to first 4 layers. In the U-Net experiments, the network was split such that first 6 layers are at the client end and the rest of the network is at the server for the label-sharing configuration. For the non-label sharing configuration, the segmentation head (consisting of the last 3 layers) is at the client-side in addition to the first 6 layers.

We do not use any form of weight synchronization; all client network segment weights are unique after training. We pass an image from a particular data source from train, validation, and test sets through the corresponding client network. For example, an image from the DT\textsubscript{5} data source, whether it be from train, validation or test set, would be passed for forward propagation through the client network residing on the client having the DT\textsubscript{5} data.

\begin{figure}
\includegraphics[width=\linewidth]{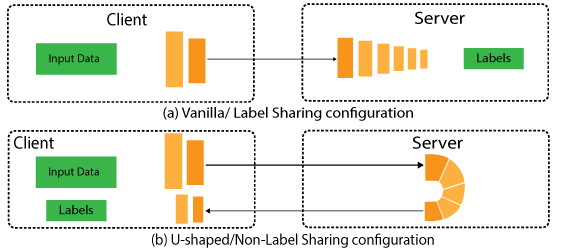}
\caption{Split Learning Configurations. In vanilla/ label sharing configuration client-side has raw input data and server-side has labels while in U-shaped/ non-label sharing configuration both the raw input data and labels are present at client-side}
\end{figure}

\subsection{SplitFed Learning Settings}
We have excluded SFLv1 from our experiments due to the unavailability of a supercomputer. We propose a novel architecture called SplitFedv3 which has the potential to outperform SL and SFLv2. As a large trainable part of the network is at the server in SL and SFLv2, “catastrophic forgetting” \cite{sheller2020federated} can happen, where the trained model favors the client data it recently used for training. In SFLv3 (as shown in Algorithm 1), client-side networks are unique for each client and the server-side network is an averaged version, the same as in SplitFedv1. The problem of catastrophic forgetting is avoided due to averaging of the server-side network. In SFLv2 and SFLv3, the split happens at the same position in the networks, as described in the split learning settings for the DenseNet and U-Net experiments. For SplitFed, we used only the alternate client training technique, and we experimented with both, the LS and the NLS configurations.

\begin{algorithm*}
\caption{SplitFedv3 algorithm for label sharing configuration. SplitFed network(W) is divided into two parts W\textsuperscript{C} and W\textsuperscript{S}. The learning rate $\eta$ is same for client-side and server-side model. Training client-side and server-side models at round t.}\label{alg:euclid}
\begin{algorithmic}[1]

\Procedure{MainServer\_Train}{}
    \Comment{$C_{\mathrm{t}}$ : set of $n_{\mathrm{t}}$ clients participating at round t}
    \For{each client $i \in C_{\mathrm{t}}$ in parallel}                \Comment{$\mathbf{W}_{i, t}^{\mathrm{C}}$ : Client-side model of client i at round t}
        \State $(\mathbf{A}_{i, t}, \mathbf{Y}_{i}) \leftarrow  \text{ClientForwardProp}(\mathbf{W}_{i, t}^{\mathrm{C}})$
        \Comment{$\mathbf{A}_{i, t}$ : Activations from client-side model of client i at round t}
        \State Pass $\mathbf{A}_{i, t}$ through $\mathbf{W}_{t}^{\mathrm{S}}$ (Forward Prop) \State Compute $\hat{\mathbf{Y}}_{i}$
        \State Loss calculation with $\mathbf{Y}_{i}$ and $\mathbf{\hat{Y}}_{i}$
        \Comment $\mathbf{Y}_{i}$ : true labels, $\mathbf{\hat{Y}}_{i}$ : predicted labels
        \State Calculate $\nabla \ell_{i}(\mathbf{W}_{t}^{S}, \mathbf{A}_{t}^{\mathbf{S}})$ (Back Prop)
        \Comment{$\nabla \ell_{i}(\mathbf{W}_{t}^{S}, \mathbf{A}_{t}^{\mathbf{S}})$ : Gradient of $\mathbf{A}_{i, t}$ }
        \State Send $d \mathbf{A}_{k, t}:=\nabla \ell_{k}(\mathbf{A}_{t}^{\mathrm{s}} ; \mathbf{W}_{t}^{\mathrm{S}})$ to client $i$
        \State ClientBackprop$(d \mathbf{A}_{i, t})$
     \EndFor{\textbf{endfor}} 
    \State Server-side model update: $\mathbf{W}_{t+1}^{\mathrm{S}} \leftarrow \mathbf{W}_{t}^{\mathrm{S}}-\eta \frac{\underline{n}_{t}}{n} \sum_{j=1}^{n} \nabla \ell_{i}(\mathbf{W}_{t}^{\mathrm{S}} ; \mathbf{A}_{t}^{\mathrm{S}})$
\EndProcedure

\State

\Procedure{ClientForwardProp}{$\mathbf{W}_{i, t}^{\mathrm{C}}$}
    \State Set $\mathbf{A}_{i, t}=\phi$
    \For{each local epoch from 1 to $E$}
    \Comment{$E$ : total number of local epochs at client end}
        \For{batch $b \in \mathcal{B}$}
        \Comment{$\mathcal{B}$ : set of local data batches}
        \State Forward propagation on $\mathbf{W}_{i, b, t}^{\mathrm{C}}$
        \State Concatenate activations from final layer of $\mathbf{W}_{i, b, t}^{\mathrm{C}}$ to $\mathbf{A}_{i, t}$
        \State Concatenate respective true labels to $Y_{i}$ 
        \EndFor{\textbf{end}}
    \EndFor{\textbf{end}}
    \State Send $\mathbf{A}_{i, t}$ and $\mathbf{Y}_{i}$ to the main server
\EndProcedure

\State

\Procedure{ClientBackprop}{$\mathbf{W}_{i, t}^{\mathrm{C}}$}
    \For{batch $b \in \mathcal{B}$}
    \State Calculate gradients $\nabla \ell_{i}(\mathbf{W}_{, b, t}^{\mathrm{C}})$ (Back Prop)
    \State $\mathbf{W}_{i, t}^{\mathrm{C}} \leftarrow \mathbf{W}_{i, t}^{\mathrm{C}}-\eta \nabla \ell_{i}(\mathbf{W}_{i, b, t}^{\mathrm{C}})$
    \EndFor{\textbf{end}}
\EndProcedure

\end{algorithmic}
\end{algorithm*}

\subsection{Evaluation Metrics}
The distributed learning techniques are evaluated on the following metrics: performance, training time, data communication, and computation. To set a benchmark for performance, we trained a model using the traditional centralized method for both sets of experiments. For evaluating performance, we use threshold diagnostic metrics:  AUROC\footnote{\url{https://en.wikipedia.org/wiki/Receiver_operating_characteristic}}, AUPRC\footnote{\url{https://machinelearningmastery.com/roc-curves-and-precision-recall-curves-for-classification-in-python/}}, and threshold-dependent techniques such as F1-score\footnote{\url{https://en.wikipedia.org/wiki/F-score}} and kappa\footnote{\url{https://en.wikipedia.org/wiki/Cohen\%27s_kappa}}. Elapsed training time, data communication, and computation are valuable metrics for distributed learning methodologies as they provide information on the feasibility of using a method in practice. We calculate all these three metrics for one epoch of model training.

\section{Results}

In this section, the performance and feasibility of distributed learning methods across various facets is evaluated and discussed.

\subsection{Performance}

No distributed learning method achieves the benchmark performance as the centralized model for the DenseNet and U-Net experiments (refer Figure 5,6,7,8,9,10,11,12 and Table 2). For DenseNet experiments (label sharing, non-label sharing, and alternate client training) and U-Net experiments (label sharing, alternate client training), SFLv3 performs better than split learning and SFLv2. Similarly, using alternate mini-batch training improves the performance of DenseNet (label sharing, non-label sharing) and U-Net (label sharing) split learning models. Further, U-Net models tend to perform better than their DenseNet counterparts. The drop in AUPRC, F1-score and, kappa is significant in DenseNet experiments. The U-Net federated learning model has the best overall performance, considering all four performance metrics.

\begin{table*}
\centering
\begin{tabular}{|l|l|l|l|l|l|l|l|l|}
\hline
Methods  & \multicolumn{8}{l|}{Performance}                         \\ \hline
 & \multicolumn{4}{l|}{DenseNet} & \multicolumn{4}{l|}{U-Net} \\ \hline
 & AUROC    & AUPRC    & F1 Score     & Kappa    &  AUROC   &  AUPRC   &  F1 Score   & Kappa  \\ \hline
Centralized &  0.9568   &  0.7629   & 0.72    & 0.69    & 0.9569    & 0.8088    & 0.75    & 0.71    \\ \hline
FL & 0.9114     & 0.652    & 0.58     & 0.52    & 0.9422    &  0.7456   & 0.74    & 0.71     \\ \hline
SL\_LS\_AC &  0.8931     & 0.5291    & 0.46    & 0.37     & 0.9282    & 0.7208    & 0.7 & 0.65 \\ \hline
SL\_LS\_AM & 0.9016    &    0.6105 & 0.5 &  0.42   &    0.9382  &  0.7322   &  0.68   & 0.64 \\ \hline
SL\_NLS\_AC &  0.872   & 0.5227     & 0.42    & 0.31    & 0.8779    & 0.6478 & 0.68  & 0.63 \\ \hline
SL\_NLS\_AM & 0.9347 & 0.7104    & 0.62    & 0.57    &  0.9036 & 0.5918 & 0.62 &    0.56 \\ \hline
SFLv2\_LS\_AC &  0.8634   & 0.5005  & 0.42    & 0.31    & 0.9146    & 0.7069 & 0.66 & 0.61    \\ \hline
SFLv2\_NLS\_AC & 0.8996  & 0.5998    & 0.53  & 0.46    &  0.8999    & 0.6912 &  0.65 & 0.6   \\ \hline
SFLv3\_LS\_AC & 0.918  & 0.6158 & 0.56 & 0.49  & 0.9319 & 0.7253 & 0.73  & 0.69    \\ \hline
SFLv3\_NLS\_AC & 0.9046 & 0.5906 & 0.55 & 0.47 & 0.9272 & 0.6314    & 0.7    & 0.66    \\ \hline
\end{tabular}
\caption{Performance of Distributed Learning methods}
\end{table*}

\begin{figure}
\includegraphics[width=\linewidth]{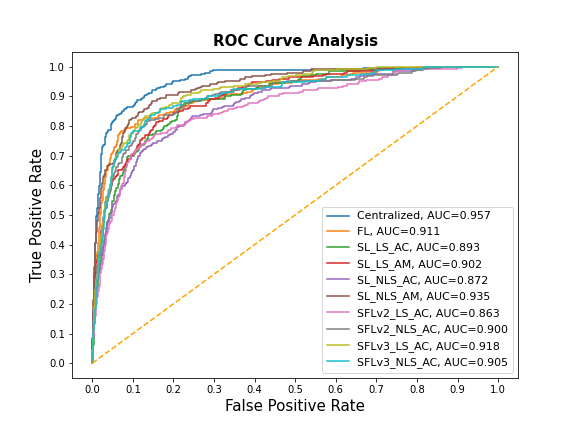}
\caption{DenseNet AUROC Curves}
\end{figure}

\begin{figure}
\includegraphics[width=\linewidth]{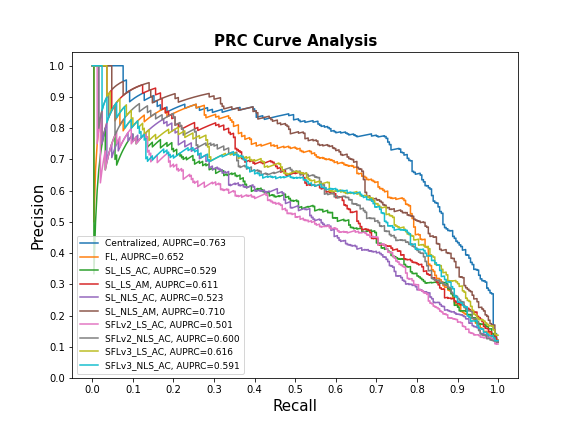}
\caption{DenseNet AUPRC Curves}
\end{figure}

\begin{figure}
\includegraphics[width=\linewidth]{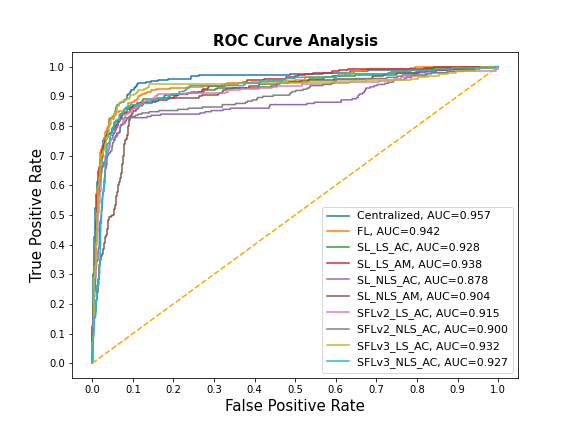}
\caption{U-Net AUROC Curves}
\end{figure}

\begin{figure}
\includegraphics[width=\linewidth]{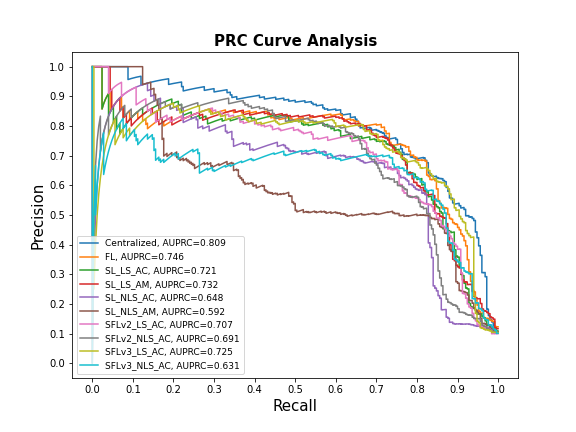}
\caption{U-Net AUPRC Curves}
\end{figure}

\begin{figure*}
\includegraphics[width=\linewidth]{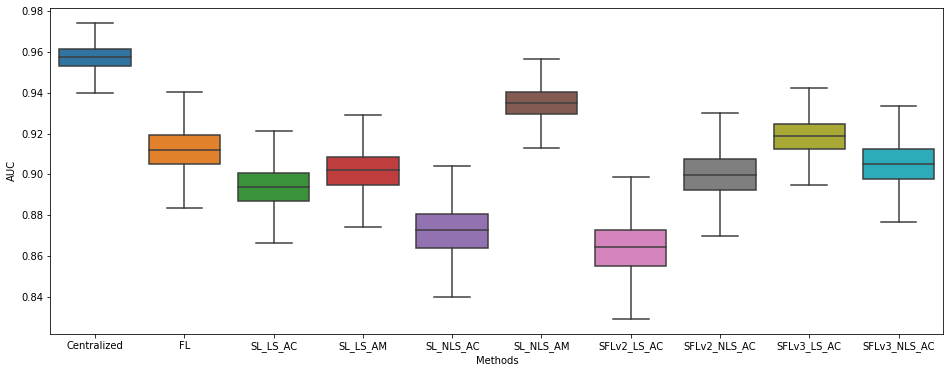}
\caption{DenseNet AUC ROC Confidence Intervals}
\end{figure*}

\begin{figure*}
\includegraphics[width=\linewidth]{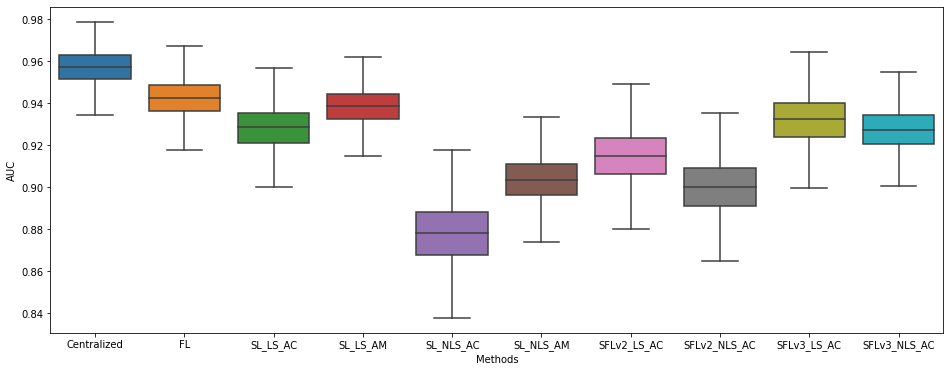}
\caption{U-Net AUC ROC Confidence Intervals}
\end{figure*}

\begin{figure*}
\includegraphics[width=\linewidth]{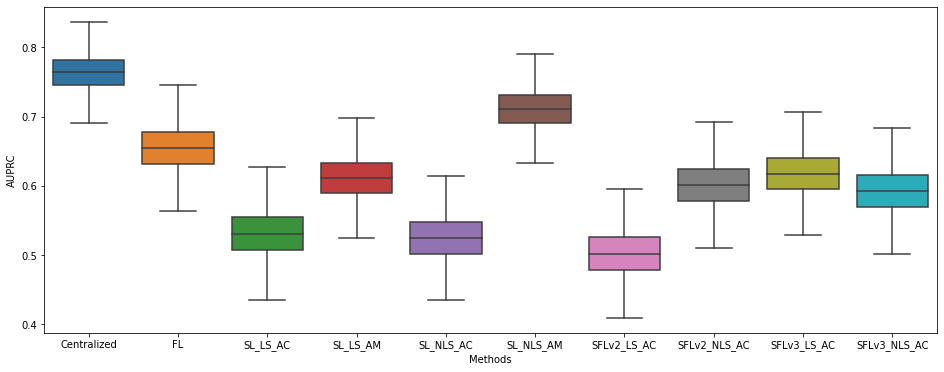}
\caption{DenseNet AUPRC Confidence Intervals}
\end{figure*}

\begin{figure*}
\includegraphics[width=\linewidth]{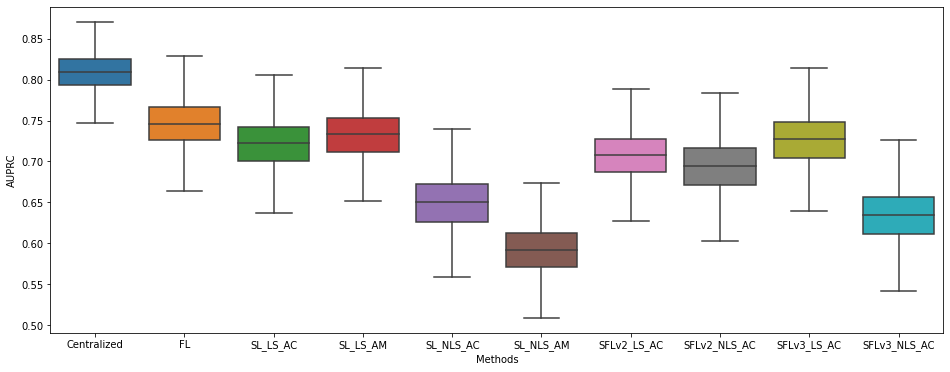}
\caption{U-Net AUPRC Confidence Intervals}
\end{figure*}

\subsection{Elapsed Training Time}
Elapsed training time is the wall clock time for training a model for 1 epoch. The time taken to train the centralized and different distributed learning models is shown in Table 3. SL, SFLv2, and SFLv3 models take almost the same time to train depending upon the configuration (label sharing or no label sharing). FL models take significantly less time to train than split learning, SFLv2, and SFLv3, for both sets of experiments.

\begin{table}[hbt!]
\centering
\resizebox{.4\textwidth}{!}{%
\begin{tabular}{|l|l|l|}
\hline
{Methods} & \multicolumn{2}{l|}{Time} \\ \cline{2-3} 
                  & DenseNet  &   U-Net          \\ \hline
Centralized       &  1 min 40 s         &  44 min 51 s         \\ \hline
FL                &  2 min 13 s         &  57 min 51 s         \\ \hline
SL\_LS\_AC        &  5 min 23 s         &  170 min 27s         \\ \hline
SL\_LS\_AM        &  5 min 22 s         &  170 min 23s           \\ \hline
SL\_NLS\_AC       &  5 min 29 s         &  279 min 17s           \\ \hline
SL\_NLS\_AM       &  5 min 42 s         &  279 min 4s           \\ \hline
SFLv2\_LS\_AC     &  5 min 24 s         & 170 min 40s          \\ \hline
SFLv2\_NLS\_AC    &  5 min 43 s          & 280 min 45s          \\ \hline
SFLv3\_LS\_AC     &  5 min 23 s          & 172 min 21s          \\ \hline
SFLv3\_NLS\_AC    &  5 min 44 s         & 279 min 24s          \\ \hline
\end{tabular}
}
\caption{Time Taken for training : 1 epoch}
\end{table}

\subsection{Data Communication}
The amount of back-and-forth data communication that takes place between the server and all clients is shown in Table 4. One epoch consists of training a model on train data and validating it on validation data for saving the weights. The data communication in federated learning consists of sending a model back and forth between the server and clients, whereas data communication for SL models consists of transfer of activations and gradients in training mode and transfer of activations in evaluation mode (validation). More data transfer occurs in non-label sharing configuration than the label-sharing configuration of SL. SFLv2 has an additional overhead of sending the client network models back-and-forth before and after averaging. Here, the client model segments are small in size (in the range of bytes) and have no significant effect on data communication for both DenseNets and U-Nets. In SFLv3, the server model segment needs to be averaged, but as it resides on the server, there is no need for transfer of the server model segment. The amount of data transfer in SL, SFLv2, and SFLv3 is enormous. Unless a strong network with high bandwidth is used, these methods seem infeasible to be used in practice. The data transfer in Federated Learning is low, which makes it suitable for use in practical settings.

\begin{table}[hbt!]
\centering
\resizebox{.4\textwidth}{!}{%
\begin{tabular}{|l|l|l|}
\hline
{Methods} & \multicolumn{2}{l|}{Data Communication} \\ \cline{2-3} 
                  & DenseNet  &   U-Net          \\ \hline
Centralized       &   -         &   -         \\ \hline
FL                &  0.13       &  0.54      \\ \hline
SL\_LS\_AC        &  14.89      &  774.05        \\ \hline
SL\_LS\_AM        &  14.89      &  774.05           \\ \hline
SL\_NLS\_AC       &  18.61      &  1474.2           \\ \hline
SL\_NLS\_AM       &  18.61      &  1474.2           \\ \hline
SFLv2\_LS\_AC     &  14.89      &  774.05          \\ \hline
SFLv2\_NLS\_AC    &  18.61      &  1474.2          \\ \hline
SFLv3\_LS\_AC     &  14.89      &  774.05          \\ \hline
SFLv3\_NLS\_AC    &  18.61      &  1474.2          \\ \hline
\end{tabular}
}
\caption{Data Communication(in GB) for 1 epoch}
\end{table}

\subsection{Computation}
The computations that occur at the server (Server Flops) and clients (Client Flops) are in the range of TeraFlops. As each client has a different number of data samples, each client would have a different number of computations. We take an average of the computations for all clients and call this measure average client flops, which is in the range of TeraFlops. In federated learning, SFLv2 and SFLv3, the server needs to average out the models. Therefore, we have included averaging model flops as an additional parameter for comparison. Averaging model flops is in the range of MegaFlops. Since an additional part of the network resides on the client in the non-label sharing configuration, it requires fewer computations than the label sharing configuration. The number of computations(Tables 5 and 6) that take place at the client is significantly greater in FL than SL. A similar number of computations happen at the clients in SL, SFLv2, and SFLv3. These distributed learning techniques leverage the splitting property to keep a large trainable part of the network at the server, drastically reducing the computations at the client end.    

\begin{table}[hbt!]
\centering
\resizebox{.45\textwidth}{!}{%
\begin{tabular}{|l|l|l|l|}
\hline
Methods &  Server &  Avg Client &  Averaging \\ \hline
Centralized & 64.21 & - &  - \\ \hline
FL & - &  12.84 &  41.73\\ \hline
SL\_LS\_AC & 61.53 & 0.53 & - \\ \hline
SL\_LS\_AM & 61.53 & 0.53 &  - \\ \hline
SL\_NLS\_AC & 61.53 & 0.53 &  - \\ \hline
SL\_NLS\_AM & 61.53 & 0.53 &  - \\ \hline
SFLv2\_LS\_AC & 61.53 & 0.53 &  0.057\\ \hline
SFLv3\_NLS\_AC & 61.53 & 0.53 &  0.069 \\ \hline
SFLv2\_LS\_AC & 61.53 & 0.53 &  41.66 \\ \hline
SFLv3\_NLS\_AC & 61.53 & 0.53 & 41.68  \\ \hline
\end{tabular}
}
\caption{Computation (Flops) for DenseNet experiments for 1 epoch. The unit for server and average client computations is TFlops and for averaging models is MFlops.}
\end{table}

\begin{table}[hbt!]
\centering
\resizebox{.45\textwidth}{!}{%
\begin{tabular}{|l|l|l|l|}
\hline
 Methods &  Server &  Avg Client &  Averaging \\ \hline
Centralized & 2129.17 & - &  - \\ \hline
FL & -  & 425.83 &  172.61\\ \hline
SL\_LS\_AC & 2064.76 & 12.83 &  - \\ \hline
SL\_LS\_AM & 2064.76 & 12.83 &  - \\ \hline
SL\_NLS\_AC & 2062.84 & 13.26 &  - \\ \hline
SL\_NLS\_AM & 2062.84 & 13.26 &  - \\ \hline
SFLv2\_LS\_AC & 2064.76 & 12.83 &  0.116\\ \hline
SFLv2\_NLS\_AC & 2062.84 & 13.26 &  0.117 \\ \hline
SFLv3\_LS\_AC &  2064.76 & 12.83 &  172.49 \\ \hline
SFLv3\_NLS\_AC & 2062.84 & 13.26 &  172.48 \\ \hline
\end{tabular}
}
\caption{Computation (Flops) for U-Net experiments for 1 epoch. The unit for server and average client computations is TFlops and for averaging models is MFlops.}
\end{table}

\section{Conclusion}

Our comparative study demonstrated the cost and feasibility of using distributed learning methods in practice. The proposed distributed learning architecture, SplitFedv3, performs better in terms of the four performance metrics (AUC, AUPRC, F1 Score, and kappa) than SL and SplitFedv2. Moreover, the new alternate mini-batch training technique improves the performance of SL models. Apart from classification performance, metrics like training time, data communication, and computational costs play a vital role in deciding the feasibility of a particular distributed deep learning method in practical settings.  The SL, SplitFedv2, and SplitFedv3 models take more time to train compared to the FL model and require more data communication. SL, SplitFedv2, and SplitFedv3 would need a high-speed network with large bandwidth to train in practical setting. However, the FL model has higher computational costs. To train an FL model, clients would require a good number of computational resources to carry out heavy computations. Unless clients have access to GPUs, the FL method would take a lot of time to carry out computations. In contrast, the clients in SL, SplitFedv2, and SplitFedv3 models would be able to carry out the small number of computations even without access to GPUs. If we take all metrics such as performance, elapsed training time, data communication and computation into account, FL is the best distributed learning method, provided clients have adequate computing power.

\bibliographystyle{ieeetr}
\nocite{*}
\bibliography{bibliography}
\end{document}